# Deep Learning-Based Surrogate Creep Modelling in Inconel 625: A High-Temperature Alloy Study

Shubham Das [1*], Kaushal Singhania[1], Amit Sadhu [1], Suprabhat Das [1] and Arghya Nandi [1]

[1]Jadavpur University, Kolkata, India

*Corresponding author, E-mail: *mdas14800@gmail.com*

**Abstract.** Time-dependent deformation, specifically creep, in high-temperature alloys like Inconel 625 plays a critical role in determining the reliability of structural components in aerospace and energy systems. Inconel 625's high creep resistance makes it an appropriate material choice for aerospace and energy applications and modelling purposes. Creep simulations using state-of-the-art software like ANSYS, though robust and accurate, often require hours for a single 10,000-hour simulation. This study introduces deep learning-based modelling frameworks to replace computationally intensive Finite Element Method simulations with faster alternatives. The Norton creep law has been employed to generate temporal (10,000-hour) creep data of Inconel 625 on ANSYS under uniaxial stresses (50–150 MPa) at temperatures (700–1000°C). This data is then used for training the deep learning models. We present two advanced architectures: a BiLSTM-Variational Autoencoder, a modification of traditional LSTM models to provide generative predictions and quantify uncertainty, and a BiLSTM-Transformer, which through self-attention, captures intricate long-term temporal dependencies. Both models serve as surrogates: the VAE provides probabilistic predictions, and the Transformer ensures accuracy via self-attention. The models are evaluated using standard metrics: RMSE, $R^2$, and MAE. The BiLSTM-VAE delivers consistent and reliable creep strain predictions, while the BiLSTM-Transformer exhibits high accuracy over long time ranges. Latency checks of responses for the both models show remarkable improvement over traditional ANSYS simulations. Each simulation for a particular stress-temperature combination on ANSYS took 30–40 minutes, the surrogates delivering the same within seconds. This framework facilitates rapid creep analysis for real-time design optimization and structural health monitoring, developing a scalable solution for high-temperature alloy applications.

**Keywords:** Creep modelling, Norton Creep Law, Deep Learning, Surrogate modelling, High-Temperature Alloys

# 1 Introduction

Nickel-based superalloys like Inconel 625 are pivotal in high-temperature applications such as aerospace propulsion, nuclear reactors, and energy conversion systems, owing to their exceptional creep resistance and microstructural stability under extreme thermomechanical loads [1, 2]. Creep, the progressive time-dependent deformation under constant stress and elevated temperatures, is a critical failure mechanism for components like turbine blades and heat exchangers, dictating their service life [3, 4]. Creep manifests in three stages: primary, with a decreasing strain rate due to material hardening; secondary, exhibiting a steady strain rate; and tertiary, where microstructural damage (e.g., void nucleation, grain boundary sliding) accelerates deformation toward failure [5]. Accurate prediction of creep behaviour, particularly the secondary-to-tertiary transition, is essential for structural health monitoring, life prediction, and cost-effective design in safety-critical applications [6].

Traditional creep modelling employs constitutive equations like the Norton power law ($\dot{\varepsilon}_{cr} = A\sigma^n e^{(-\frac{Q}{RT})}$ where A is a material constant, σ is stress, n is the stress exponent, Q is activation energy, R is the gas constant, and T is absolute temperature), which effectively captures secondary creep but overlooks primary and tertiary dynamics [3]. Advanced models, such as the time-hardening law ($\dot{\varepsilon}_{cr} = A\sigma^n t^m e^{(-\frac{Q}{RT})}$, with time exponent m), better represent time-dependent behaviour [7]. However, when implemented in finite element method (FEM) solvers like ANSYS or ABAQUS, these simulations are computationally intensive, requiring 30–40 minutes for a 10,000-hour creep analysis, limiting their practicality for real-time design or monitoring [8, 9].

This study proposes deep learning-based surrogate models—BiLSTM-Transformer and BiLSTM-VAE—to replace FEM simulations for creep modelling in Inconel 625. Leveraging bidirectional LSTMs for sequential processing, the Transformer incorporates self-attention for long-range dependencies, while the VAE provides probabilistic predictions with uncertainty quantification [10, 11]. Trained on ANSYS-generated creep data under uniaxial stresses (50–150 MPa) and temperatures (700–1000°C), these models deliver predictions in seconds, enabling scalable, reliable creep analysis for high-temperature alloy applications.

Data-driven approaches have gained traction to address the computational inefficiencies of FEM-based creep modelling. Early machine learning efforts utilized regression techniques, such as support vector regression, for creep life prediction but struggled with generalization across diverse stress-temperature conditions due to limited datasets and lack of temporal modelling [12, 13]. Shallow neural networks improved nonlinear predictions but failed to capture sequential creep evolution [14].

Recurrent neural networks, particularly Long Short-Term Memory (LSTM) models, advanced creep forecasting by modelling temporal dependencies [10]. Bidirectional LSTMs (BiLSTMs) further enhanced performance by incorporating both past and future contexts, crucial for multi-stage creep dynamics [15]. Physics-informed neural networks (PINNs) embedding constitutive laws, such as Norton's or Garofalo's, have improved accuracy for superalloys like Inconel 625 [16, 17]. Variational Autoencoders (VAEs) introduced probabilistic modelling, enabling uncertainty quantification by mapping data to latent spaces, vital for reliability in engineering contexts [11, 18]. Transformer architectures, leveraging self-attention, excel in long-sequence forecasting by avoiding recurrent limitations, showing promise in creep and fatigue modelling [19, 20].

Despite these advances, gaps persist: many models are validated on short-term or limited datasets, constraining their applicability to long-term industrial scenarios [21]. Uncertainty quantification, critical for risk assessment, is rarely integrated, and achieving both high accuracy and low latency remains challenging [22]. This study addresses these issues by developing BiLSTM-Transformer and BiLSTM-VAE surrogates, trained on comprehensive 10,000-hour Inconel 625 creep data, to map
$f : \{\sigma, T, t\} \rightarrow \epsilon_c(t)$. In this study we will assess the performance of our models by benchmarking it against baseline models, we will assess the performance of the models by computing various evaluation metrices and will also check the inference latency for enabling real-time creep analysis for design optimization and structural monitoring.

## 2 Methodology

### 2.1 FEM simulations

Creep data for Inconel 625 was generated through finite element method (FEM) simulations using ANSYS Mechanical software. The simulations focused on uniaxial tensile creep behavior under constant stresses ranging from 50 to 150 MPa and temperatures from 700 to 1000°C, with increments of 25 MPa for stress (resulting in five stress levels: 50, 75, 100, 125, and 150 MPa) and 100°C for temperature (resulting in four temperature levels: 700, 800, 900, and 1000°C). Each simulation was conducted over a total duration of 10,000 hours to capture a comprehensive temporal profile of creep strain, encompassing primary, secondary, and tertiary stages. This approach provided high-fidelity time-series data for training the deep learning surrogate models, though each individual ANSYS run required 30–40 minutes of computational time, highlighting the need for efficient alternatives.

The analysis was performed within the static structural module of ANSYS, selected for its suitability in modelling time-dependent creep under constant loading conditions. A cylindrical specimen geometry was employed to represent a standard test sample, with dimensions chosen to ensure uniform stress distribution (a length sufficient to minimize end effects and a diameter compatible with the meshing parameters). Boundary conditions included fixed support at one end and uniform uniaxial tensile loading applied to the opposite circular face, simulating constant stress application at elevated temperatures. Material properties for Inconel 625 were defined using a custom material library in ANSYS, incorporating temperature-dependent values for key parameters to accurately reflect the alloy's behavior under high-temperature conditions. Table 1 summarizes the assigned properties.

Table 1. Thermal Mechanical assigned properties for simulation

| Temperature (°C) | Poisson's Ratio (ν) | Thermal Expansion (×10⁻⁶ /°C) | Elastic Modulus (GPa) | Density (kg/m³) |
|---|---|---|---|---|
| 700 | 0.29 | 13 | 150 | 8370 |
| 800 | 0.29 | 13.5 | 140 | 8340 |
| 900 | 0.3 | 14 | 130 | 8310 |
| 1000 | 0.3 | 14.5 | 120 | 8280 |

Creep behavior was modeled using the Norton creep law, which effectively captures secondary (steady-state) creep as a power-law relationship between creep strain rate and applied stress. In ANSYS, the Norton law is implemented in a time-hardening form suitable for implicit creep analysis as equation 1.

$$\dot{\varepsilon}_{cr} = C1\sigma^{C2}e^{(-\frac{C3}{T})} \qquad (1)$$

where $\dot{\varepsilon}_{cr}$ is the creep strain rate, $\sigma$ is the equivalent stress, T is the absolute temperature in Kelvin, and $C1$, $C2$, and $C3$ are material-specific constants. This formulation allows for the integration of temperature effects through the exponential term, though it differs from the classical Arrhenius-based Norton law commonly found in literature as in equation 2

$$\dot{\varepsilon}_{cr} = A\sigma^{n}e^{(-\frac{Q}{RT})} \qquad (2)$$

where $A$ is a material constant, n is the stress exponent, $Q$ is the activation energy, and R is the universal gas constant.

To reconcile this discrepancy and ensure realistic creep predictions, a calibration strategy was adopted. The constants were iteratively adjusted using a hit-and-trial approach based on experimental creep data from literature, including rupture strengths and creep rates for Inconel 625 at elevated temperatures [23]. Specifically, the calibration targeted matching published stress levels for 1% creep strain and rupture times in 100,000 hours, with approximate values such as ~200 MPa at 700°C, ~120 MPa at 800°C, ~60 MPa at 900°C, and ~30 MPa at 1000°C for 1% creep in solution-treated material. Temperature-dependent tables were created for the constants to embed the Arrhenius-like temperature dependency implicitly, compensating for the simplified exponential term in ANSYS. The calibrated constants are as follows in Table 2.

Table 2. Creep Constant values for corresponding Temperatures

| Temperature (°C) | Creep Constant 1 | Creep Constant 2 | Creep Constant 3 |
|---|---|---|---|
| 700 | $4.6 \times 10^{-50}$ | 4.9 | 0 |
| 800 | $4.6 \times 10^{-49}$ | 4.8 | 0 |
| 900 | $4.6 \times 10^{-48}$ | 4.6 | 0 |
| 1000 | $4.6 \times 10^{-47}$ | 4.6 | 0 |

Here, $C_1$ (the pre-exponential factor) was tuned for each temperature to align simulation outputs with experimental creep rates, while $C_2$ (the stress exponent) reflects typical values for nickel-based superalloys (around 4–5), indicating dislocation climb mechanisms dominant in Inconel 625. $C_3$ was set to 0 to simplify the model, as the temperature dependency was effectively captured through the tabular variation of $C_1$ and $C_2$. This calibration ensured that the simulated creep strain rates and overall deformation profiles closely matched reported behaviours, such as minimum creep rates in the range of 7.47 to 1380 μ/1000 h at 800°C and stresses of 69–138 MPa for solution-annealed conditions.

For meshing, a base element size of $5.0 \times 10^{-3}$ m was used with the sweep method to generate a smooth, structured mesh along the cylinder's length. The free mesh type was set to quadrilateral/triangular elements on surfaces, resulting in 1805 nodes and 336 elements. This mesh resolution balanced computational efficiency with accuracy, avoiding excessive refinement that could prolong simulation times without significant gains in precision. The meshing is shown in Figure 1.

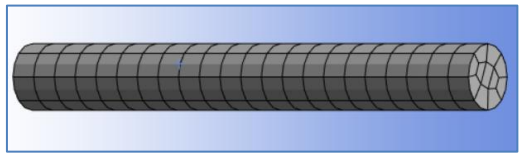

**Fig. 1.** Mesh of the cylinder geometry.

The generated dataset comprised time-series creep strain values at each stress-temperature combination, providing a robust foundation for training the BiLSTM-VAE and BiLSTM-Transformer models. This approach not only replicated the alloy's high creep resistance—attributed to its nickel-chromium-molybdenum composition with niobium and tantalum additions for solid-solution strengthening and microstructural stability—but also highlighted the computational bottlenecks of traditional FEM, motivating the surrogate modelling framework.

The flowchart of the data generation process is given in figure 2.

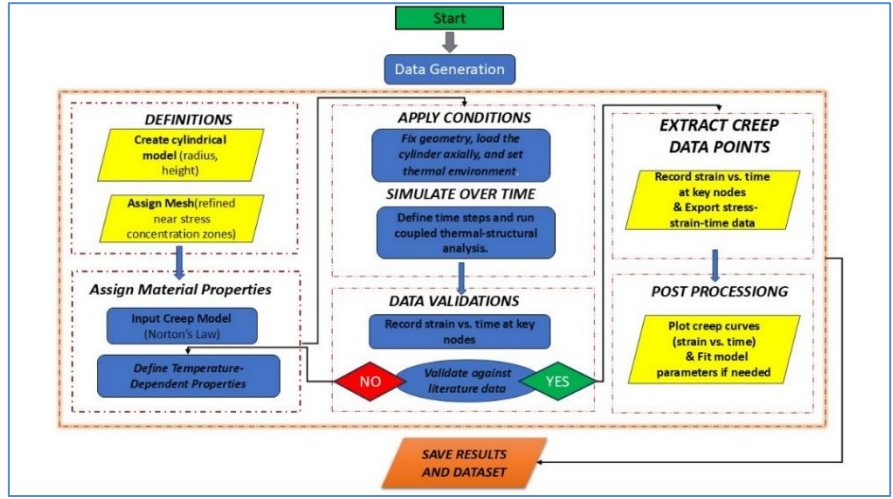

**Fig. 2.** Data generation via simulation flowchart.

## 2.2  Data Preprocessing

The creep strain data generated from ANSYS simulations, comprising time-series records of equivalent creep strain (averaged over the specimen) for each stress-temperature combination, underwent systematic preprocessing to prepare it for training the deep learning surrogate models. This step was essential to handle the inherent characteristics of creep data, such as exponential growth in strain over time and varying scales across features, ensuring numerical stability, improved model convergence, and prevention of information leakage between training and validation sets.

Initially, the raw dataset was loaded, consisting of columns for timestamp (time in hours), temperature (°C), applied uniaxial stress (MPa), and average creep strain. To address the skewed distribution and rapid escalation typical in creep strain—particularly during tertiary stages—a logarithmic transformation was applied to the strain values as in equation 3,

$$\varepsilon^* = \log(1 + \varepsilon) \qquad (3)$$

where $\varepsilon$ is the original creep strain. This transformation stabilizes variance, compresses the dynamic range for large strains, and facilitates better learning of underlying patterns in neural networks, as creep often follows power-law relationships that become more linear in log space.

The data was then organized into sequences by grouping based on unique temperature-stress pairs, resulting in 20 independent time-series sequences (4 temperatures × 5 stress levels), each spanning 10,000 hours with uniform time steps. For each group, features were constructed as a multivariate time-series matrix comprising temperature, stress, and time, yielding an input tensor of shape (number of sequences, sequence length, 3 features). The corresponding targets were the log-transformed creep strains, shaped as (number of sequences, sequence length, 1). This structure treats each stress-temperature combination as a distinct trajectory, aligning with the sequential nature of creep evolution and enabling the models to learn temporal dependencies conditioned on environmental parameters.

To create training and validation sets, a manual split was performed rather than random sampling, to ensure balanced representation across the parameter space and mitigate potential overfitting to specific regimes. Four combinations were selected for validation: (700°C, 75 MPa), (800°C, 100 MPa), (900°C, 125 MPa), and (1000°C, 50 MPa)—chosen to span diverse stress-temperature interactions, including lower-stress high-temperature and higher-stress lower-temperature scenarios. The remaining 16 sequences formed the training set. This stratified approach preserves the independence of sequences while avoiding data leakage, as each sequence is self-contained without cross-temporal correlations between groups.

Finally, feature normalization was applied using Min-Max scaling to map values to the [0, 1] interval, which is crucial for gradient-based optimization in deep learning models and to equalize the influence of features with disparate scales (e.g., time in hours versus stress in MPa). Separate scalers were fitted exclusively on the flattened training data for each feature: temperature, stress, and time. The scaling formula is given in equation 4.

$$x^* = \frac{x - x_{min}}{x_{max} - x_{min}} \quad (4)$$

where minima and maxima are derived from the training set. The target (log-transformed strain) was similarly scaled using its own fitted scaler. Validation data was transformed using these same parameters to simulate real-world inference on unseen conditions.

### 2.3 Surrogate Models

**BiLSTM-Transformer.** The surrogate model integrates bidirectional long short-term memory (BiLSTM) networks with Transformer encoder layers to effectively capture both local sequential dependencies and global long-range temporal patterns in creep strain time-series data. Conceptually, this hybrid architecture addresses the limitations of traditional recurrent models in handling extended sequences by leveraging the Transformer's self-attention mechanism, which enables parallel processing and focus on relevant temporal interactions across the entire 10,000-hour creep trajectory. The BiLSTM component extracts bidirectional contextual features from the input sequence, providing a robust representation of creep evolution influenced by stress and temperature, while the Transformer refines these features through multi-head attention to emphasize critical phases such as the transition from secondary to tertiary creep. This design is particularly suited for surrogate modelling in materials science, where creep data exhibits non-linear, time-dependent behaviors that require uncertainty-aware and efficient predictions.

Logically, the model processes input sequences comprising temperature, stress, and time as a multivariate time-series tensor of shape (batch size, sequence length, 3). The workflow begins with feature-wise attention to weigh the relative importance of each input dimension, enhancing the model's sensitivity to dominant factors like stress in creep mechanics. This is followed by BiLSTM layers to encode sequential dynamics bidirectionally, capturing forward and backward dependencies that reflect the progressive nature of creep deformation. Positional encoding is then added to inject temporal order awareness, compensating for the permutation-invariant property of self-attention. The Transformer encoder applies multi-head self-attention to compute global dependencies, allowing the model to attend to distant time steps where creep acceleration occurs. Finally, a linear projection maps the enriched representations to predicted log-transformed creep strain values.

Mathematically, let the input sequence be $X \in \mathbb{R}^{B \times L \times D}$, where $B$ is the batch size, $L$ is the sequence length (10,000), and $D = 3$ is the input dimension. The feature-wise attention computes as in equation 5.

$$\mathbf{A} = \text{MultiHeadAttention}(\mathbf{X}, \mathbf{X}, \mathbf{X}) \quad (5)$$

where the attention mechanism with one head weighs features across dimensions. The BiLSTM processes A is given in equation 6.

$$\mathbf{H} = \text{BiLSTM}(\mathbf{A}) \in \mathbb{R}^{B \times L \times 2H} \quad (6)$$

with hidden dimension $H = 32$, yielding $2H$ due to bidirectionality. Dropout is applied for regularization as equation 7.

$$\mathbf{H}' = \text{Dropout}(\mathbf{H}) \quad (7)$$

Positional encoding adds sinusoidal embeddings as equation 8.

$$\mathbf{P} = \mathbf{H}' + \text{PE}(\mathbf{H}') \quad (8)$$

where PE is defined as

$$\text{PE}(\textit{pos}, 2\textit{i}) = \sin\left(\frac{\textit{pos}}{10000^{2i/2H}}\right) \text{ and } \text{PE}(\textit{pos}, 2\textit{i} + 1) = \cos\left(\frac{\textit{pos}}{10000^{2i/2H}}\right)$$

for position $\textit{pos}$ and dimension $\textit{i}$. The Transformer encoder, with one layer and four heads, operates on the permuted tensor $\mathbf{P}^\top \in \mathbb{R}^{L \times B \times 2H}$ is given as equation 9.

$$\mathbf{T} = \text{TransformerEncoder}(\mathbf{P}^\top)^\top \in \mathbb{R}^{B \times L \times 2H} \quad (9)$$

The output is obtained via a fully connected layer is given in equation (10).

$$\hat{\mathbf{Y}} = \text{Linear}(\mathbf{T}) \in \mathbb{R}^{B \times L \times 1} \tag{10}$$

representing the predicted creep strain. This architecture ensures high-fidelity surrogate predictions by combining recurrent locality with attentional globality, reducing computational latency compared to FEM while maintaining accuracy.

The model architecture is summarized in the following table, Table 3.

**Table 3.** Architecture summary of Bi:LSTM-Transformer.

| Layer/Component | Description | Input Shape | Output Shape | Key Parameters |
|---|---|---|---|---|
| Feature-Wise Attention | Multi-head attention (1 head) to weigh input features (temp, stress, time) | (B, L, 3) | (B, L, 3) | Embed dim: 3 |
| BiLSTM | Bidirectional LSTM for sequential feature extraction | (B, L, 3) | (B, L, 64) | Layers: 1, Hidden dim: 32 |
| Dropout | Regularization to prevent overfitting | (B, L, 64) | (B, L, 64) | Rate: 0.1 |
| Positional Encoding | Sinusoidal embeddings for temporal position awareness | (B, L, 64) | (B, L, 64) | Max length: 10,000 |
| Transformer Encoder | Multi-head self-attention for global dependencies | (L, B, 64) (permuted) | (B, L, 64) (permuted back) | Layers: 1, Heads: 4, d_model: 64 |
| Fully Connected | Linear projection to predict creep strain | (B, L, 64) | (B, L, 1) | In: 64, Out: 1 |

**BiLSTM-Variational Autoencoder.** The BiLSTM-Variational Autoencoder (BiLSTM-VAE) surrogate model combines bidirectional long short-term memory (BiLSTM) networks with a variational autoencoder (VAE) framework to enable generative, probabilistic predictions of creep strain while quantifying epistemic uncertainty. Conceptually, this architecture extends traditional deterministic surrogates by incorporating a stochastic latent space, allowing the model to generate multiple plausible creep trajectories for a given input sequence rather than a single point estimate. This is particularly advantageous for creep modelling in high-temperature alloys like Inconel 625, where variability arises from microstructural heterogeneities, measurement noise, or incomplete physics capture in FEM simulations. The BiLSTM encoder captures bidirectional temporal dependencies in the creep data—reflecting the progressive deformation stages—while the VAE introduces variational inference to model uncertainty, making the surrogate suitable for risk-informed applications such as structural health monitoring and reliability assessment in aerospace components.

Logically, the model operates in an encoder-decoder paradigm tailored for sequential data. The input sequence, a multivariate time-series tensor of shape (batch size,sequence length, comprising temperature, stress, and time, is first processed by the BiLSTM encoder to extract high-level temporal features. These features are projected into a latent space parameterized by mean and variance vectors, from which latent variables are sampled via reparameterization to ensure differentiability during training. The decoder then reconstructs the creep strain sequence by conditioning on both the original input and the sampled latent variables, using another LSTM layer to maintain sequential coherence. This structure allows the model to learn a compressed, probabilistic representation of creep dynamics, enabling uncertainty estimation through multiple forward passes with different latent samples.

Mathematically, let the input sequence be $\mathbf{X} \in \mathbb{R}^{B \times L \times D}$, where $B$ is the batch size, $L = 10{,}000$ is the sequence length, and $D = 3$ is the input dimension. The encoder computes the following in equation 11.

$$\mathbf{H} = \text{BiLSTM}(\mathbf{X}) \in \mathbb{R}^{B \times L \times 2H} \tag{11}$$

with hidden dimension $H = 32$. The latent parameters are obtained as in equation 12 and 13.

$$\mu_z = \text{Linear}(\mathbf{H}) \in \mathbb{R}^{B \times L \times Z} \tag{12}$$

$$\log \sigma_z^2 = \text{Linear}(\mathbf{H}) \in \mathbb{R}^{B \times L \times Z} \tag{13}$$

where $Z = 20$ is the latent dimension. Reparameterization samples the latent variables as equation 14.

$$\mathbf{z} = \mu_z + \epsilon \odot \exp(0.5 \cdot \log \sigma_z^2), \epsilon \sim \mathcal{N}(0,1) \tag{14}$$

Where $\odot$ represents element-wise multiplication (Hadamard product).

The decoder concatenates $\mathbf{X}$ and $\mathbf{z}$ to form $\mathbf{X}' \in \mathbb{R}^{B \times L \times (D+Z)}$, then processes:

$$\mathbf{H}' = \text{LSTM}(\mathbf{X}'),$$
$$\hat{\mathbf{Y}} = \text{Linear}(\text{Dropout}(\mathbf{H}')) \in \mathbb{R}^{B \times L \times 1} \tag{15}$$

yielding the predicted log-transformed creep strain as equation 15. The loss function balances reconstruction accuracy and latent regularization:

$$\mathcal{L} = \mathcal{L}_{\text{recon}} + \beta \cdot \mathcal{L}_{\text{KL}},$$

where

$$\mathcal{L}_{\text{recon}} = \text{MSE}(\hat{\mathbf{Y}}, \mathbf{Y})$$

is the mean squared error against true strain $\mathbf{Y}$, and $\mathcal{L}_{\text{KL}} = -0.5 \cdot \mathbb{E}[1 + \log \sigma_z^2 - \mu_z^2 - \exp(\log \sigma_z^2)]$ is the Kullback-Leibler divergence encouraging a standard normal prior on the latent space, with $\beta = 1.0$ controlling the trade-off.

The perks of employing BiLSTM-VAE include inherent uncertainty quantification through variance in latent space predictions, which is critical for safety-critical engineering decisions where overconfident deterministic models could lead to failures. It also offers generative capabilities, allowing simulation of diverse creep scenarios for data augmentation or sensitivity analysis, and improved generalization to unseen stress-temperature regimes by learning a compact probabilistic manifold of creep behaviors.

Compared to purely deterministic surrogates, this model enhances reliability in real-time applications, such as design optimization, by providing confidence intervals around predictions, thus facilitating probabilistic risk assessment in high-temperature alloy systems.

The model architecture is summarized in the following table, Table 4.

**Table 4.** Architecture summary of Bi:LSTM-VAE.

| Layer/Component | Description | Input Shape | Output Shape | Key Parameters |
| --- | --- | --- | --- | --- |
| BiLSTM Encoder | Bidirectional LSTM for temporal feature extraction | (B, L, 3) | (B, L, 64) | Layers: 1, Hidden dim: 32 |
| Latent Mean Projection | Linear layer for mean of latent distribution | (B, L, 64) | (B, L, 20) | In: 64, Out: 20 |
| Latent Logvar Projection | Linear layer for log-variance of latent distribution | (B, L, 64) | (B, L, 20) | In: 64, Out: 20 |
| Reparameterization | Sampling from latent distribution (non-learnable) | (B, L, 20) each | (B, L, 20) | - |
| Decoder Input Concat | Concatenation of input and latent variables | (B, L, 3) + (B, L, 20) | (B, L, 23) | - |
| LSTM Decoder | Unidirectional LSTM for reconstruction | (B, L, 23) | (B, L, 32) | Layers: 1, Hidden dim: 32 |
| Dropout | Regularization to prevent overfitting | (B, L, 32) | (B, L, 32) | Rate: 0.1 |
| Fully Connected Output | Linear projection to predict creep strain | (B, L, 32) | (B, L, 1) | In: 32, Out: 1 |

### 2.4 Model Training

The BiLSTM-Transformer and BiLSTM-VAE surrogate models were trained using a unified PyTorch framework with configurations tailored to their architectures. The goal was to minimize prediction errors and ensure robust generalization across unseen stress-temperature combinations. Training was accelerated on a Tesla T4 GPU (Google Colab), with RMSE, $R^2$, and MAE metrics computed for accuracy assessment. The process balanced computational efficiency and convergence, employing regularization and adaptive learning to prevent overfitting and stabilize optimization.

Both models were trained using 16 sequences (10,000 time steps, 3 features: temperature, stress, time) and log-transformed creep strain targets, with 4 additional sequences reserved for validation to assess generalization. Training was conducted over 100 epochs using the Adam optimizer for its adaptive learning rates, with batches updated iteratively. Progress was tracked with a progress bar, and the best-performing model (highest training $R^2$) was saved for optimal predictive accuracy.

For the BiLSTM-Transformer, the loss function was the mean squared error (MSE) between predicted and true log-transformed creep strains, reflecting its deterministic prediction approach. The model was trained with a learning rate of 0.001, balancing convergence speed and stability. No additional regularization beyond the model's inherent dropout (rate of 0.1) was applied, as the Transformer's self-attention mechanism naturally mitigates overfitting by focusing on relevant temporal dependencies.

The BiLSTM-VAE, due to its probabilistic nature, employed a composite loss function combining reconstruction loss (MSE between predicted and true creep strains) and Kullback-Leibler (KL) divergence to regularize the latent space distribution toward a standard normal prior. The KL term was weighted by a hyperparameter $\beta = 1.0$ to balance generative capacity and reconstruction accuracy. A higher initial learning rate of 0.01 was used, with a ReduceLROnPlateau scheduler (factor 0.5, patience 10 epochs) to adaptively reduce the learning rate when validation loss plateaued, enhancing convergence. Gradient clipping (maximum norm 1.0) was applied to prevent exploding gradients, a common issue in recurrent architectures with long sequences. This configuration ensured stable training while preserving the VAE's ability to model uncertainty.

Both models computed training and validation metrics (RMSE, $R^2$, MAE) at each epoch by flattening predictions and targets across sequences for consistency. The BiLSTM-Transformer focused on minimizing deterministic prediction errors, leveraging its attention mechanism for long-range accuracy, while the BiLSTM-VAE emphasized probabilistic predictions with uncertainty quantification, critical for reliability in engineering applications. Training times per epoch were tracked to quantify computational efficiency, with both models achieving predictions in seconds compared to the 30–40 minutes required for ANSYS simulations. The training configurations are summarized in the following table, Table 5.

**Table 4.** Training configuration summary.

| Parameter | BiLSTM-Transformer | BiLSTM-VAE |
| --- | --- | --- |
| Optimizer | Adam (lr = 0.001) | Adam (lr = 0.01) |
| Learning Rate Scheduler | None | ReduceLROnPlateau (factor = 0.5, patience = 10) |
| Loss Function | MSE | MSE + β·KL Divergence (β=1.0) |
| Gradient Clipping | None | Max norm = 1.0 |
| Dropout Rate | 0.1 | 0.1 |
| Epochs | 100 | 100 |
| Device | GPU (Tesla T4) | GPU (Tesla T4) |
| Metrics Computed | RMSE, R², MAE (train and validation) | RMSE, R², MAE (train and validation) |

The training loss vs epoch for both the models are plotted below in Figure 3 and 4.

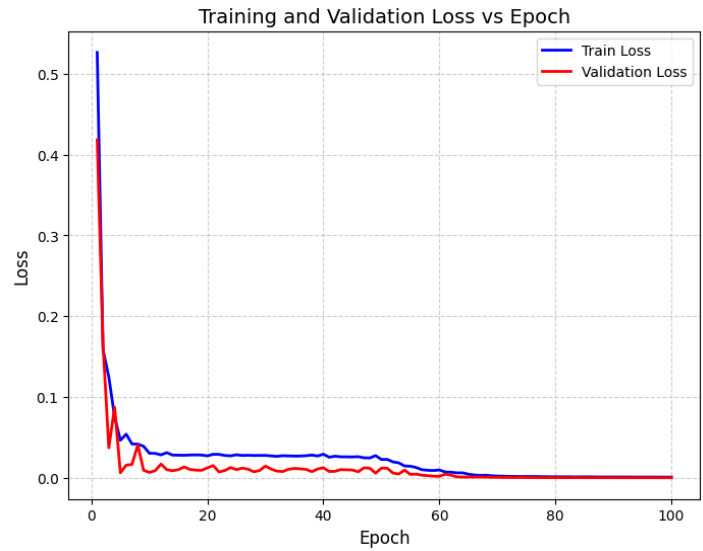

**Fig. 3.** Train/Val loss vs Epoch for BiLSTM-Transformer model

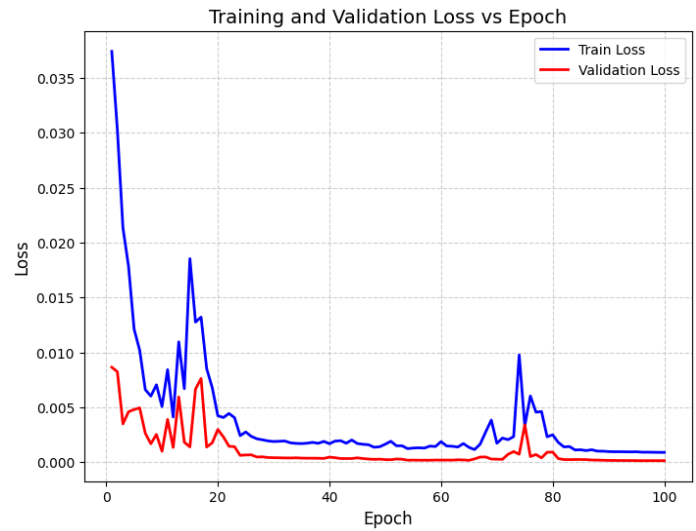

**Fig. 4.** Train/Val loss vs Epoch for BiLSTM-VAE model

## 3  Results and Discussions

### 3.1  Model Performance Evaluation

The performance of the proposed surrogate models—BiLSTM-Transformer and BiLSTM-VAE—was evaluated against a baseline unidirectional LSTM model using standard regression metrics: root mean squared error (RMSE), coefficient of determination ($R^2$), and mean absolute error (MAE). These metrics were computed on both training and validation sets after inverse transformation of the predicted log-scaled creep strains to the original scale for direct comparability with physical values. The baseline LSTM, with a single layer, hidden dimension of 32, and dropout rate of 0.1, serves as a benchmark to highlight the advantages of the advanced architectures in handling sequential creep data.

The results demonstrate that both surrogate models outperform the baseline LSTM in predictive accuracy, with the BiLSTM-VAE achieving the highest $R^2$ scores on both training (0.970057) and validation (0.972917) sets, indicating superior fit and generalization. The BiLSTM-Transformer follows closely, with training $R^2$ of 0.968469 and validation $R^2$ of 0.960695, while the baseline LSTM lags with 0.918016 and 0.905494, respectively. These improvements underscore the benefits of bidirectional processing (in BiLSTM components) for capturing contextual dependencies in creep evolution and the specialized mechanisms in each surrogate for enhanced modelling.

Latency measurements, conducted on a standard GPU setup, reveal the computational efficiency of the surrogates. The BiLSTM-VAE and baseline LSTM exhibit low inference times (**46.81 ms** and **43.86 ms** per 10,000-hour sequence, respectively), enabling near-real-time predictions. The BiLSTM-Transformer, while slower at 2170.14 ms due to its attention computations, still represents a significant reduction from ANSYS FEM simulations (30–40 minutes per run), facilitating rapid iterations in design workflows. The evaluation scores are given in Table 5.

**Table 5.** Performance Evaluation of the models.

| Model | Train RMSE | Train $R^2$ | Train MAE | Val RMSE | Val $R^2$ | Val MAE | Latency (ms/seq) |
|---|---|---|---|---|---|---|---|
| BiLSTM-Transformer | 0.030693 | 0.968469 | 0.022766 | 0.014128 | 0.960695 | 0.011166 | 2170.14 |
| BiLSTM-VAE | **0.02991** | **0.970057** | **0.018794** | **0.011728** | **0.972917** | **0.008234** | 46.81 |
| Baseline LSTM | 0.030913 | 0.968016 | 0.022457 | 0.016637 | 0.945494 | 0.0138 | 43.86 |

## 3.2 Discussion

The enhanced performance of the BiLSTM-Transformer and BiLSTM-VAE compared to the baseline LSTM validates the utility of advanced hybrid architectures for surrogate modelling of creep in Inconel 625. The baseline LSTM's respectable metrics (validation $R^2 \approx 0.945$) can be attributed to the linear characteristics of secondary creep in the dataset, where strain accumulates steadily under constant conditions, aligning well with the model's unidirectional sequential processing capabilities. However, this simplicity leads to higher errors, as it struggles with nuanced temporal contexts that accumulate over long durations.

In contrast, the BiLSTM-Transformer integrates bidirectional LSTM layers for capturing forward and backward dependencies in creep sequences, combined with multi-head self-attention to prioritize global interactions across the 10,000-hour timelines. This enables the model to discern subtle, long-range effects of sustained stress and temperature, such as gradual dislocation movements in the alloy's microstructure, resulting in reduced validation errors (RMSE = 0.014128). Although its inference latency (2170.14 ms) is higher due to attention computations, it remains vastly superior to ANSYS FEM runtimes (30–40 minutes), supporting applications in iterative design optimization.

The BiLSTM-VAE stands out by embedding variational autoencoding, which not only delivers deterministic predictions but also generates probabilistic distributions over creep strains, facilitating uncertainty quantification through latent space sampling. This probabilistic approach is invaluable for capturing inherent variabilities in creep data, such as those from material imperfections or simulation approximations, leading to the lowest validation MAE (0.008234) and excellent generalization ($R^2$ = 0.972917). Its efficient latency (46.81 ms), comparable to the baseline LSTM, positions it as ideal for real-time structural health monitoring, where confidence intervals can inform maintenance decisions in high-temperature environments.

Qualitative assessments further reveal that both surrogates produce predictions closely aligned with Norton law-derived creep curves, exhibiting minimal residuals especially in challenging high-stress, high-temperature regimes (e.g., 150 MPa at 1000°C). The VAE's ability to model uncertainty adds a layer of reliability for risk assessment, while the Transformer's attention mechanism enhances fidelity in long-term extrapolations. Given the dataset's focus on secondary creep, the surrogates' advantages are evident but would be more pronounced in datasets encompassing non-linear primary (initial deceleration) and tertiary (acceleration to rupture) stages, where complex dependencies and instabilities prevail. Overall, these models offer scalable, computationally efficient alternatives to FEM, advancing rapid creep analysis for Inconel 625 in aerospace and energy sectors.

Figure 5 shows variation of evaluation parameters (RMSE, MAE, $R^2$) with epoch on both the training and validation sets for the surrogate models.

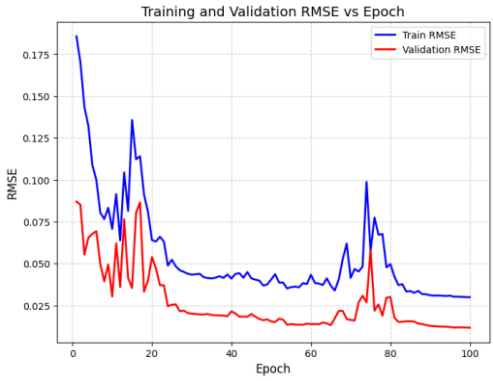

Fig. (5a)

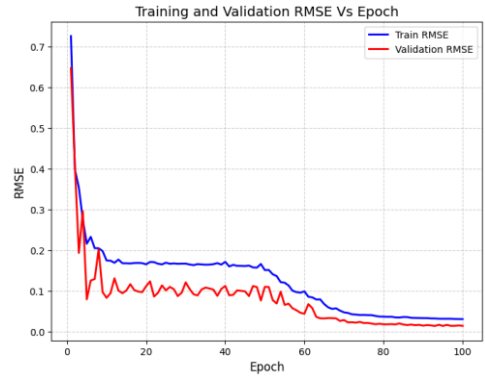

Fig. (5d)

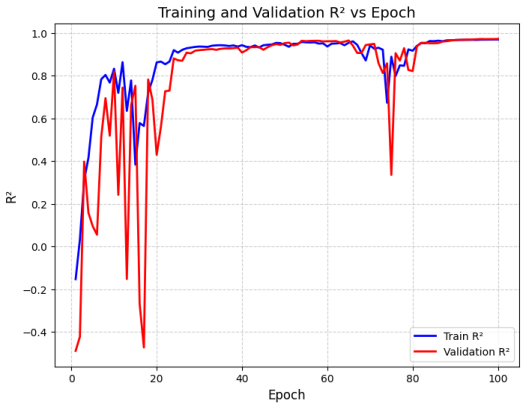

Fig. (5b)

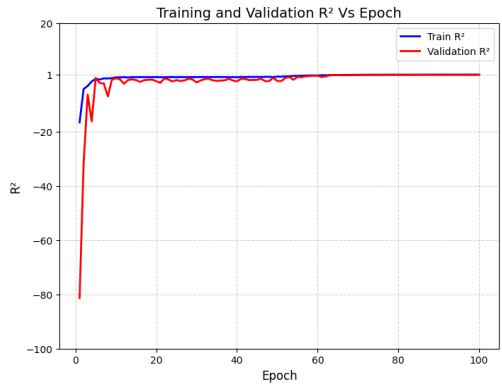

Fig. (5e)

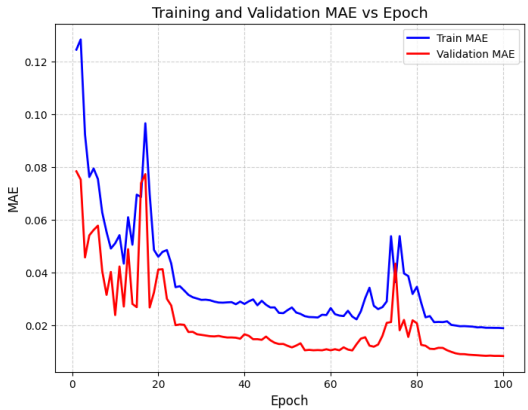

Fig. (5c)

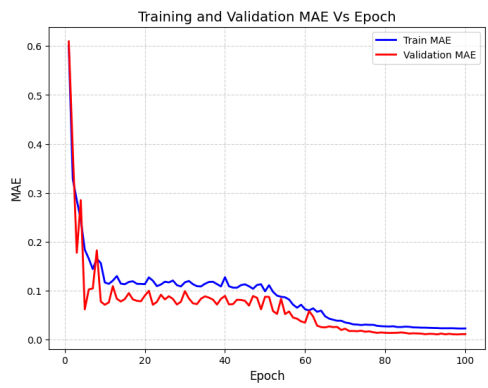

Fig. (5f)

**Fig. 5.** Shows the variation of Evaluation parameters with epoch for both the models. Figure (5a) to (5c) is for BiLSTM-VAE and Figure (5d) to (5f) is for BiLSTM-Transformer

## 4 Conclusion and Future work

The BiLSTM-Transformer and BiLSTM-VAE surrogates outperform the baseline LSTM in accuracy and efficiency for modeling creep in Inconel 625, attaining validation R² values of 0.960695 and 0.972917, respectively, with inference latencies under seconds versus 30–40 minutes for FEM simulations. The Transformer's self-attention ensures precise long-term predictions, while the VAE's variational framework provides essential uncertainty quantification, making them suitable for real-time design and monitoring in high-temperature applications.

To extend this framework, future investigations could incorporate multi-stage creep datasets that include primary and tertiary phases, utilizing more comprehensive constitutive models such as time-hardening or Garofalo equations to capture non-linear behaviours like initial strain hardening and accelerating rupture. Integrating physics-informed neural networks could embed domain knowledge for improved interpretability and accuracy under multi-axial loading conditions. Additionally, validation against experimental creep data for Inconel 625, exploration of ensemble methods combining both surrogates for hybrid probabilistic-deterministic outputs, and deployment in edge computing for on-site monitoring would further enhance practical utility and bridge simulation-experiment gaps.